\documentclass[10pt,twocolumn,letterpaper]{article}

\usepackage[latin9]{inputenc}
\usepackage{xcolor}
\usepackage{array}
\usepackage{float}
\usepackage{multirow}
\usepackage{amsmath}
\usepackage{amssymb}
\usepackage{graphicx}
\PassOptionsToPackage{normalem}{ulem}
\usepackage{ulem}
\usepackage[unicode=true,
 bookmarks=false,
 breaklinks=true,pdfborder={0 0 1},backref=section,colorlinks=false]
 {hyperref}
\hypersetup{
 pagebackref=true,letterpaper=true,colorlinks}

\makeatletter

\usepackage{iccv}
\usepackage{times}
\usepackage{epsfig}
\usepackage{graphicx}
\usepackage{algorithm}  
\usepackage{algpseudocode}  
\usepackage{amsmath}
\usepackage{amssymb}
\usepackage{verbatim}

\usepackage[breaklinks=true,bookmarks=false]{hyperref}

\iccvfinalcopy 

\def\iccvPaperID{3258} 

\ificcvfinal\fi

\begin{document}

\title{Hierarchical Object-to-Zone Graph for Object Navigation}

\author{Sixian Zhang$^{1,2}$, Xinhang Song$^{1,2}$\footnotemark[1], Yubing Bai$^{1,2}$, Weijie Li$^{1,2}$, Yakui Chu$^{4}$, Shuqiang Jiang$^{1,2,3}$\\
\small \textsuperscript{1}Key Lab of Intelligent Information Processing Laboratory of the Chinese Academy of Sciences (CAS), \\
\small Institute of Computing Technology, Beijing \textsuperscript{2}University of Chinese Academy of Sciences, Beijing \\
\small \textsuperscript{3}Institute of Intelligent Computing
Technology, Suzhou, CAS \textsuperscript{4}Huawei Application Innovate Laboratory, Beijing \\
{\tt\small \{sixian.zhang, xinhang.song, yubing.bai, weijie.li\}@vipl.ict.ac.cn }\\
{\tt\small chuyakui@huawei.com; sqjiang@ict.ac.cn}
}

\maketitle
\ificcvfinal\thispagestyle{empty}\fi

\renewcommand{\thefootnote}{\fnsymbol{footnote}}{
\footnotetext[1]{Corresponding author.}}

\begin{abstract}
The goal of object navigation is to reach the expected objects according
to visual information in the unseen environments. Previous works usually
implement deep models to train an agent to predict actions in real-time.
However, in the unseen environment, when the target object is not
in egocentric view, the agent may not be able to make wise decisions
due to the lack of guidance. In this paper, we propose a hierarchical
object-to-zone (HOZ) graph to guide the agent in a coarse-to-fine
manner, and an online-learning mechanism is also proposed to update
HOZ according to the real-time observation in new environments. In
particular, the HOZ graph is composed of scene nodes, zone nodes and
object nodes.  With the pre-learned HOZ graph, the real-time observation
and the target goal, the agent can constantly plan an optimal path from
zone to zone. In the estimated path, the next potential zone is regarded
as sub-goal, which is also fed into the deep reinforcement learning
model for action prediction. Our methods are evaluated on the AI2-Thor
simulator. In addition to widely used evaluation metrics SR and SPL, we also propose a new evaluation metric of SAE that focuses on the effective action rate. Experimental results demonstrate the effectiveness and efficiency of our proposed method.
The code is available at \url{https://github.com/sx-zhang/HOZ.git}.
\end{abstract}

\section{Introduction}

\begin{figure}
\begin{centering}
\includegraphics[scale=0.3]{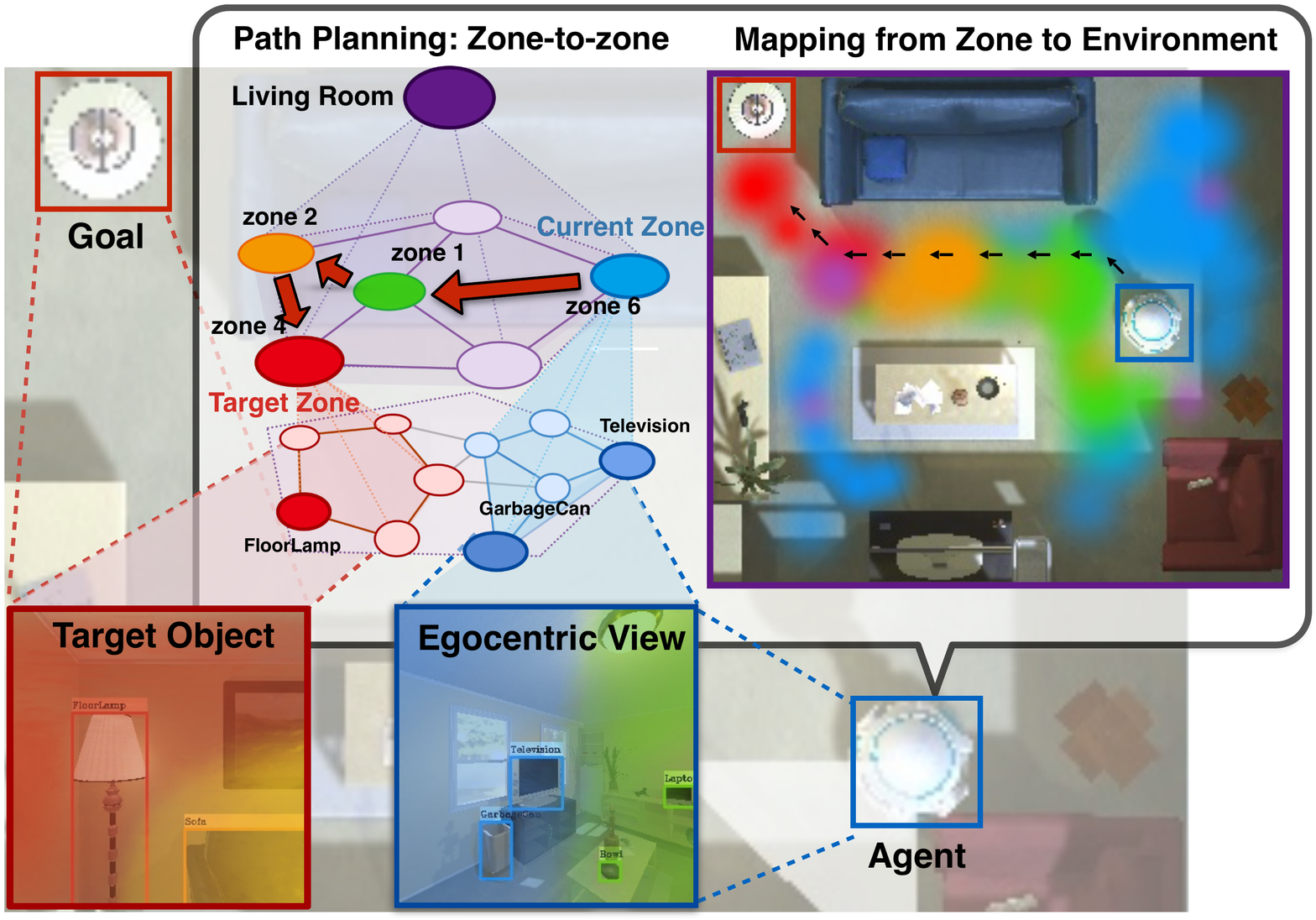}
\par\end{centering}
\caption{\label{fig:motivation}Overview of object navigation with HOZ graph.
At the beginning, agent locates at the Current Zone ($zone_{6}$,
blue) and the goal \textit{FloorLamp} belongs to Target Zone ($zone_{4}$,
red). The HOZ graph plans a real-time optimal path ($zone_{6}-zone_{1}-zone_{2}-zone_{4}$).
Then agent's next sub-goal is $zone_{1}$ (green). In the same
way, the agent keeps updating sub-goal until it arrives at the target.
Note that each color implies specific location and direction where
agent can observe similar views.}

\vspace{-10pt}
\end{figure}

Visual navigation task requires the agent to reach a specified goal.
Conventional methods usually require spatial layout information, such
as maps of the environments, which can be easily obtained in seen
environments while unavailable in unseen environments. Therefore,
how to efficiently navigate to the target in unseen environments is
typically challenging.

With the visual input of egocentric observation, previous works \cite{Mnih_2015_Nature,Piotr_ICLR17,A3C}
learn a deep reinforcement learning policy by maximizing the reward.
The key challenge in those works is the generalization to unseen environments
\cite{savn}, especially when the target is not in the sight. Therefore,
more recent works \cite{scene_priors,ECCV_relation_graph} attempt
to embed prior knowledge, such as object graph and relation graph,
to improve the navigation model's generalization ability. In particular,
Yang \textit{et al.} \cite{scene_priors} construct an object-to-object
graph, which provides correlated objects as auxiliary information
to locate the target object. Their object graph is too general to
fit into specific environments. Additionally, Du \textit{et al.} \cite{ECCV_relation_graph}
propose to learn object relation graph, which fits the testing environments
better than the general object graph. The above approaches focus on
constructing object-oriented graph to provide some clues to the navigation
when the target is not in the view. However, since object relations
and spatial layout are usually inconsistent in different environments,
the generalization ability of the above methods are still limited.

Motivated by enhancing the generalization ability of the navigation
model, we carry out this study from two aspects: 1) learning an adaptive
spatial knowledge representation that is applicable to various environments;
2) adapting the learned knowledge to guide navigation in the unseen
environments. Besides, regions in larger area are considered in our
knowledge, which are denoted as zones. Compared with objects,
larger zones are more likely to be observed by agent. Thus our core
idea for navigation guidance is zone.

In this paper we propose the hierarchical object-to-zone (HOZ)
graph to capture the prior knowledge of scene layout for object navigation
(see Figure \ref{fig:motivation}). During training, we construct a
general HOZ graph from all scenes, as rooms in the same scene category
have same spatial structures. Each scene node corresponds to a scene-wise
HOZ graph, whose zone nodes are obtained by matching and merging the
room-wise HOZ graphs. For each room-wise HOZ graph, each zone node
represents a group of relevant objects and each zone edge models the
adjacent probability of two zones. Then we train a zone-to-action
LSTM policy via deep reinforcement learning in the photo-realistic
simulator AI2-Thor \cite{ai2thor}. For each episode, the pre-learned
HOZ graph helps to plan an optimal path from current zone to target
zone, thus deducing the next potential zone on the path as a sub-goal.
The sub-goal is embedded with graph convolutional network (GCN) to
predict actions. Considering that different environments have diverse
zone layouts, we also propose an online-learning mechanism to update
the general learned HOZ graph according to current unseen environment.
In this way, the initial HOZ graph will evolve towards current environment\textquoteright s
specific layout and help agent to navigate successfully. Note that
the update only holds for an episode and each episode starts from
the initial HOZ graph. In addition to widely used evaluation metrics
Success Rate (SR) and Success weighted by Path Length (SPL), we also
propose a new evaluation metric of Success weighted by Action Efficiency
(SAE) that considers the efficiency of the navigation action into
SR. Our experiments show that the HOZ graph outperforms the baseline by a large margin. In summary, our contributions are as follows:
\begin{itemize}
\item We propose to learn the hierarchical object-to-zone (HOZ) graph that
captures prior knowledge to guide object navigation agent with easier
sub-goals.
\item We propose a new evaluation metric named Success weighted by Action
Efficiency (SAE).
\item By integrating HOZ graph into a zone-to-action policy, the navigation
performance can be significantly improved in SR, SPL and SAE metrics.
\end{itemize}

\begin{figure*}
\noindent \begin{centering}
\includegraphics[scale=0.7]{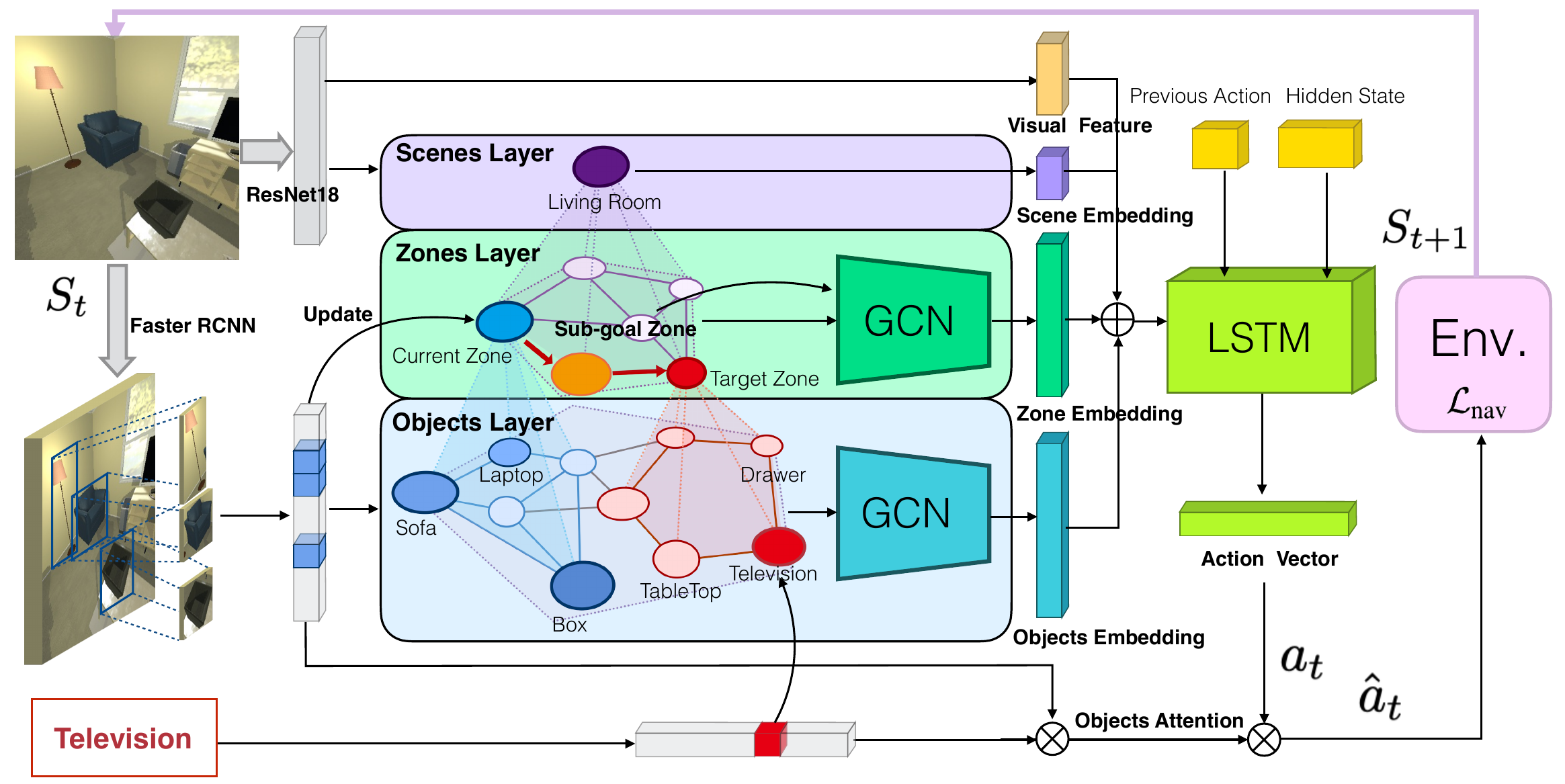}
\par\end{centering}
\caption{\label{fig:framework}\textbf{Model Overview.} Our model is composed
of the hierarchical object-to-zone (HOZ) graph and the zone-to-action
LSTM. Given the target object and current observations, the agent
first recognizes the scene category, locates the current zone, and
deduces the next sub-goal zone according to the HOZ graph. The HOZ
graph is updated at each timestamp based on the observations of
the unseen environment. The zone-to-action LSTM learns to predict
efficient actions based on the concatenated information provided by
the HOZ graph.}

\vspace{-10pt}
\end{figure*}

\section{Related Work}

\textbf{Geometry-based navigation:} Conventional navigation methods
typically use a map as reference, whether it is constructed in advance
or built simultaneously during visual navigation. \cite{1_Kidono,17_Singh}
utilize the metric-based map to perceive the environment and \cite{19_Elfes}
keeps updating a probabilistic chessboard representation during agent\textquoteright s
locomotion. Comparatively, \cite{7_Savinov,24_Chen,36_Chaplot} adopt
coarse-grained topological map, with nodes showing semantic features
and edges reasoning spatial relationships. \cite{2_Thrun,21_Thrun}
both integrate metric-based map and topological map to improve mobile
robot navigation. \cite{Linegar_ICRA} constructs an experience graph
to deal with long-term appearance changes. In addition, \cite{5_Gupta}
adopts a belief map as spatial memory. Rather than relying on a specific
map, our HOZ graph acts as prior knowledge to aid navigation in unseen
environments.

\textbf{Learning-based navigation:} Deep learning has gained popularity
in end-to-end localization, exploration and so on \cite{5_Gupta,7_Savinov}.
As an early try, \cite{20_Meng} takes neural networks to build a hallway
follower model in indoor navigation. Nowadays, many researches turn
to\textbf{ }reinforcement learning (RL) to help agents make action
decisions \cite{15_Sadeghi,17_Singh,8_Kahn}. To improve generalization,
\cite{Target-driven_navigation,scene_priors,Wu_iccv19} all employ
Actor-Critic model \cite{A3C}. Moreover, \cite{18_Chen} learns exploration
policies using an intrinsic coverage reward in imitation learning.
\cite{33_Li} trains a task generator and a meta-learner to learn transferable
meta-skills. \cite{3_Cummins} uses a generative model with probabilistic
framework to benefit the similarity calculation of two observations.
\cite{7_Savinov,22_Bansal} propose a waypoint navigation to find
simpler sub-goals. \cite{9_Mousavian} utilizes semantic information
to boost deeper understanding. Meanwhile, \cite{23_Fang} puts forward
a memory-based policy. They embed each observation into a memory and
perform this spatial-temporal memory on three visual navigation tasks.
\cite{Meng_2020_ICRA} proposes a reachability estimator that provides
the navigator a sequence of target observations to follow. This line
of works mostly treat the policy network as a black box and train
it via RL, whereas our HOZ graph includes coarse-to-fine inputs of
object, region, and scene, which allows for interpretable navigation.

\textbf{Goal-driven navigation:} This kind of navigation is carried
out for subjective purposes, mainly conducted by natural language
instructions or target images. It can be distinguished into PointGoal
navigation \cite{5_Gupta,17_Singh} and ObjectGoal navigation \cite{Piotr_ICLR17,23_Fang,savn,9_Mousavian,scene_priors,Wu_iccv19}.
In particular, sometimes the target may be presented as an image \cite{36_Chaplot,Target-driven_navigation}.
Our work focuses on object navigation in unseen indoor environments.
\cite{savn} proposes a self-adaptive visual navigation method to help
agent learn to learn in an unseen environment via meta-reinforcement
learning. \cite{ECCV_relation_graph} proposes an object representation
graph to learn the spatial correlations among different object categories,
and uses imitation learning to train the agent. A memory-augmented
tentative policy network is used to detect deadlock conditions and
provides additional action guidance during testing. Recent works have
applied knowledge graphs to image classification \cite{Marino_2017_CVPR},
segmentation \cite{Zhu_2015_CVPR}, zero-shot recognition \cite{Wang_2018_CVPR}
and navigation \cite{scene_priors,Wu_iccv19}. \cite{Wu_iccv19} proposes
Bayesian Relational Memory that captures the room-to-room prior layout
of environments during training to produce sub-goals for semantic-goal
visual navigation. \cite{scene_priors} establishes an object-to-object
graph by extracting the relationships among object categories in Visual
Genome \cite{Visual_Genome}. While in our work, we conduct the online-learning
hierarchical object-to-zone (HOZ) graph to serve as prior knowledge
for object navigation, which provides more general regional information.

\section{Preliminary Notation\label{sec:Preliminary-Notation}}

Considering a set of environments $Q$ and objects $P$, in each navigation
episode, agent is initialized to a random location $l=\left\{ x,z,\theta_{yaw},\theta_{pitch}\right\} $
in an environment $q\in Q$. $x,z$ represent the plane coordinate
and $\theta_{yaw},\theta_{pitch}$ represent the yaw and pitch angle
(of the agent). At each timestamp $t$, agent learns a policy function
$\pi\left(a_{t}|o_{t},p\right)$, which predicts an action $a_{t}\in\mathcal{A}$
based on first-person view $o_{t}$ and the target object $p\in P$.
The discrete action space $\mathcal{A}=\{ MoveAhead,RotateLeft, RotateRight, LookDown,\\ LookUp, Done \}$.
Note that the action $Done$ is judged by the agent itself rather
than informed by the environment. The success of object navigation
task requires agent finally capturing and getting close to the target
object (less than a threshold).

\section{Hierarchical Object-to-Zone (HOZ) Graph}

Our goal is to navigate agent to the given target without a precise
map in the unseen environment. Thus, a great challenge in such task
is to locate objects. Previous works \cite{ECCV_relation_graph,savn,scene_priors} directly take
the target object embedding as the goal to guide action prediction. However, it's typically difficult to plan an efficient
path without prior knowledge about the unknown environment. The agent
in those works might not find the path at the beginning, leading to
some meaningless actions, such as frequently spinning around and backing.
In order to provide stronger guidance, our navigation model considers
a wider range region where the target object may be located, which
is denoted as zone.

Each zone usually consists of a group of relevant objects. For instance,
microwave, cooker and sink usually appear in the same zone. Thus,
navigating to microwave may first require locating such zone. Since precise  map information is not available in the unseen
environment, how to collect suitable zones information and construct
a hierarchical object-to-zone (HOZ) graph remains challenging. Therefore,
we start from seen scenes to construct HOZ graph (Section \ref{subsec:HOZ-Graph-Construction})
and later adaptively update it when navigating in the unseen scenes
(Section \ref{subsec:HOZ-Graph-Embedding}). 

We consider the zones from the following hierarchical
structure. Our environments consist of several scenes, such as bedroom,
living room, and kitchen, etc, and each scene contains several rooms.
In each room $i\in\left\{ 1,2,\ldots,n\right\} $ , we get room-wise
HOZ graph $\Omega_{i}\left(V_{i},E_{i}\right)$, whose zone nodes
are obtained by clustering the egocentric observation features and
edges are defined as the adjacent probability of two zones (traced
back to co-occurrence probability of each contained objects). Then
we fuse these room-wise HOZ graphs grouped by scene to obtain scene-wise
HOZ graphs $G_{s}\left(V_{s},E_{s}\right)$. All scene-wise HOZ graphs
have the same structure and constitute our final HOZ graph (Section
\ref{subsec:HOZ-Graph-Construction}).

\subsection{HOZ Graph Construction\label{subsec:HOZ-Graph-Construction}}

\begin{algorithm}[t]
\caption{Scene-wise HOZ graph construction\label{alg:Category-wise-HOZ-graph}}

\begin{algorithmic}[1]
	\Require 			
		$K$: zone number 
	\Require 
		$(Room_{1}, \ldots, Room_{n})$ of same scene category
	\State Create room-wise HOZ graphs set $\Omega$
	\For {$i$ $\gets$ $1$ to $n$}
		\State Get features and locations $[(f_{1},l_{1}),\cdots ,(f_{d},l_{d})]$ 
		\Statex \qquad \qquad in  $Room_{i}$ by agent with random exploring
		\State Create a graph $G_{r}(V_{r}, E_{r})$
		\State $(C_{1}, \cdots, C_{K}) \gets$ K-Means$(f_{1},\cdots,f_{d}, K)$ 			
		\State $V_{r} \gets$ cluster centers $(C_{1}, \cdots, C_{K})$
		\State $E_{r} \gets$ calculate edges with Equation \ref {eq:edges}
		\State Add room-wise HOZ graph to $\Omega_{i} \gets G_{r}(V_{r}, E_{r})$
	\EndFor
	\State Create scene-wise HOZ graph $G_{s}(V_{s}, E_{s}) $
	\State Initialize $G_{s}(V_{s}, E_{s}) \gets \Omega_{1}$
	\For {$i$ $\gets$ $2$ to $n$}
		\State Create  weighted bipartite graph $G^{b}(V^{b},E^{b})$
		\State $V^{b} \gets V_{s}$ (all nodes of $G_{s}$), $V_{i}$ (all nodes of $\Omega_{i}$)
		\State $\omega(E^{b}) \gets$ calculate similarity by Equation \ref{eq:weight} 
		\State Perfect matching $\varPsi^{*} \gets$ Kuhn-Munkres( $\omega(E^{b})$ )
		\State Update $G_{s} \gets Avg(G_{s}, \Omega_{i}, \varPsi^{*})$ refer to Figure \ref{fig:Matching-and-merging.}
	\EndFor
	\Ensure scene-wise HOZ graph $G_{s}(V_{s}, E_{s})$ 
\end{algorithmic}
\end{algorithm}

\subsubsection{Room-wise HOZ graph}

Similar scenes (e.g. ``living room'') may consist of common objects
and object layouts \cite{cvpr_hamid,cvpr_zhen}. For instance, when
referring to the living room, an area composed of sofa, pillow and
table, or an area composed of TV set and TV cabinet may appear in
our mind. When searching for an object, humans tend to first locate
the typical area where the object most likely to appear. In our work,
we denote such areas as zones and embed zones to
guide agent. In order to obtain those representative
zones, we sample visual features around the room and make a clustering
on them.

In a specific room $i$, we first let the agent explore the room to collect
a set of visual tuple features $\left(f,l\right)$, where $f\in\mathbb{R}^{N\times1}$
is a bag-of-objects vector obtained by Faster-RCNN \cite{Faster_R-CNN},
representing the objects that appear in the current view. It should
be noticed that we use the bag-of-objects vector composed of 0 and
1 to represent the object category. If the current view contains many objects belonging to the same category, we only record them once. $N$ denotes the number of object
categories, and $l=\left\{ x,z,\theta_{yaw},\theta_{pitch}\right\} $
denotes the observation location defined in Section \ref{sec:Preliminary-Notation}.
Then we make K-Means clustering on features $f$ to get $K$ zones,
forming the zone nodes in room-wise HOZ graph $\Omega_{i}\left(V_{i},E_{i}\right)$.
We use $v_{k}$ and $\delta\left(v_{k}\right)$ to represent the $k$-th
zone node and its embedded feature. The embedded feature represents
the cluster center, which is calculated by $\delta\left(v_{k}\right)=\frac{1}{|zone_{k}|}\sum_{\left(f_{\gamma},l_{\gamma}\right)\in zone_{k}}f_{\gamma}$,
where $zone_{k}$ is a group of clustered visual tuple features $\left(f,l\right)$
after K-Means, and $|zone_{k}|$ is the element number. Each dimension's value of $\delta\left(v_{k}\right)$ shows the connection
relationship between the zones layer and objects layer (Figure \ref{fig:framework}),
representing the co-occurrence frequency of objects belonging to the
$zone_{k}$.

The edge $e\left(v_{k},v_{j}\right)$ in the zones layer, represents
the probability that two zones are adjacent to each other, which can
be calculated as follows:

\begin{equation}
\begin{array}{c}
e\left(v_{k},v_{j}\right)=\frac{\sum_{\left(f_{\gamma},l_{\gamma}\right)\in zone_{k}}\sum_{\left(f_{\zeta},l_{\zeta}\right)\in zone_{j}}\eta\left(l_{\gamma},l_{\zeta}\right)}{|zone_{k}|\times|zone_{j}|}\\
\eta\left(l_{\gamma},l_{\zeta}\right)=\begin{cases}
1 & \left|x_{\gamma}-x_{\zeta}\right|+\left|y_{\gamma}-y_{\zeta}\right|\leq\varepsilon\\
0 & otherwise
\end{cases}
\end{array}\label{eq:edges}
\end{equation}

where $\varepsilon$ is a hyper-parameter threshold. Then we use all
node features to recognize the scene category. For each room $i$,
we construct a room-wise HOZ graph $\Omega_{i}\left(V_{i},E_{i}\right)$.

\subsubsection{Scene-wise HOZ graph}

\begin{figure}[t]
\begin{centering}
\includegraphics[scale=0.24]{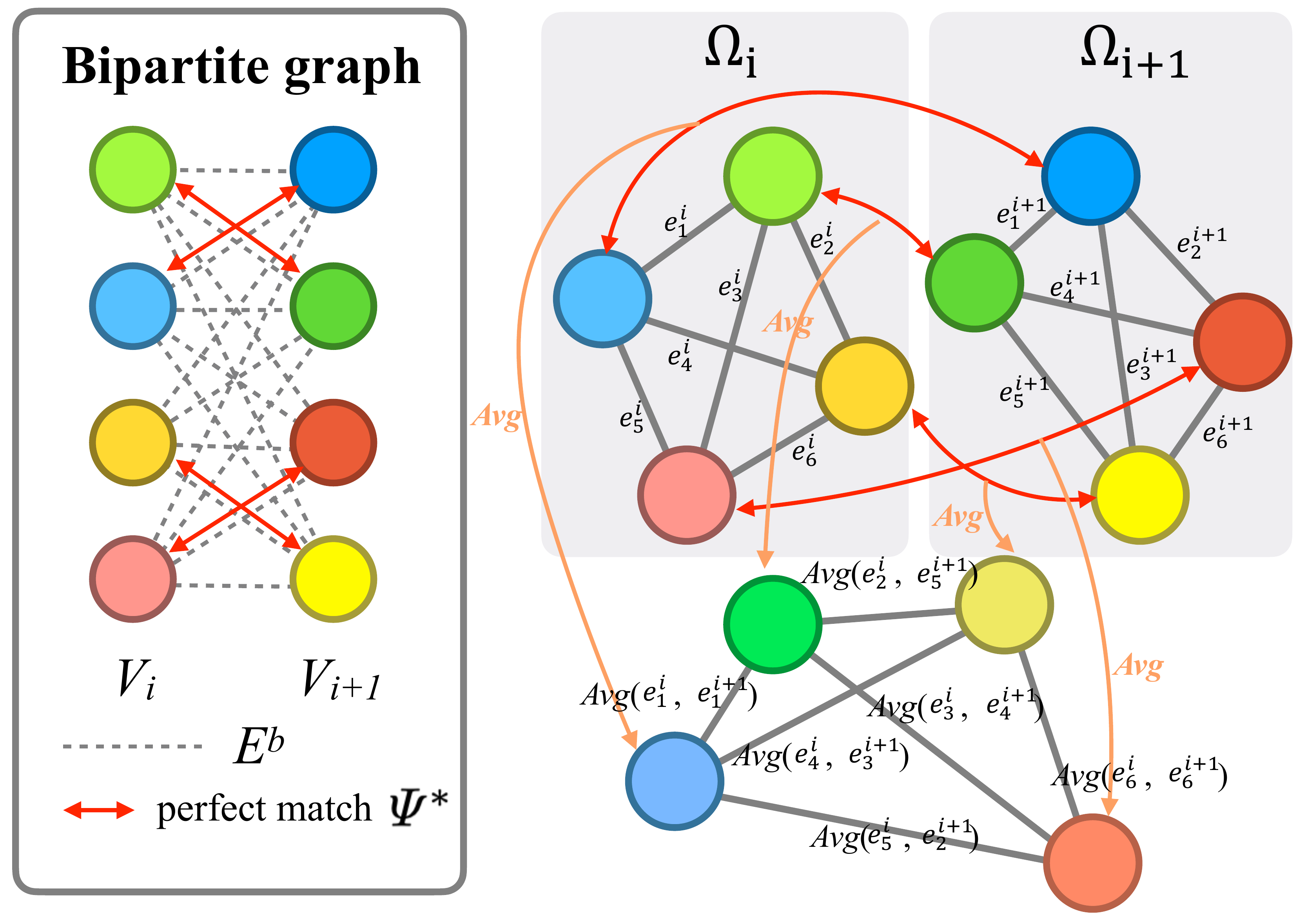}
\par\end{centering}
\caption{\textbf{Matching and Merging. }The left part shows the perfect maximum
matching on weighted bipartite graph with Kuhn--Munkres algorithm.
The right part shows the average calculation of merging two matched
room-wise HOZ graphs into a new graph. For instance, two nodes (in
red) are matched, and merged with average pooling (written as $Avg$). Correspondingly, edges between these nodes are merged with average
pooling. \label{fig:Matching-and-merging.}}

\vspace{-10pt}
\end{figure}

To obtain scene-wise HOZ graph, we group all room-wise HOZ graphs
by scene category. Take one scene as an example, we can obtain the room-wise
set $\Omega=\left\{ \Omega_{1}\left(V_{1},E_{1}\right),\ldots,\Omega_{n}\left(V_{n},E_{n}\right)\right\} $.
Since the zones number $K$ is fixed, each room-wise HOZ graph
has the same structure for later matching and merging. Considering
that directly computing the maximum matching of all room-wise HOZ
nodes is expensive, we propose pairwise perfect matching and merging
on two graphs each time until all graphs merge into the final one.
The matching between $\Omega_{i}\left(V_{i},E_{i}\right)$ and $\Omega_{i+1}\left(V_{i+1},E_{i+1}\right)$
graphs can be regarded as the weighted bipartite graph matching. We
construct a bipartite graph $G^{b}=\left(V_{i}\cup V_{i+1},E^{b}\right)$,
where $V_{i}$ is the nodes set in $\Omega_{i}$, $|V_{i}|=|V_{i+1}|$,
and $E^{b}$ represents all fully connected edges. A perfect matching
is to find a subset $\varPsi\subseteq E^{b}$, where each node has
exactly one edge incident on it. The maximum perfect matching satisfies
$\varPsi^{*}=\underset{\Psi}{argmax}\sum_{e^{b}\in\varPsi}\omega\left(e^{b}\right)$,
where $e^{b}\equiv e^{b}\left(v_{k},v_{j}\right)$ represents the
edge matching nodes $v_{k}$ and $v_{j}$, $v_{k}\in V_{i}$, $v_{j}\in V_{i+1}$.
The weight function $\omega(e^{b})$ calculates the similarity of two nodes as
\begin{equation}
\omega\left(e^{b}\left(v_{k},v_{j}\right)\right)={1}/{d\left(\delta_{k},\delta_{j}\right)}\label{eq:weight}
\end{equation}

and $d\left(\delta_{k},\delta_{j}\right)$ is defined as

\begin{equation}
d\left(\delta_{k},\delta_{j}\right)=\sqrt{\left(\delta_{k}-\delta_{j}\right)^{T}\left(\delta_{k}-\delta_{j}\right)}+\frac{1}{\delta_{k}^{T}\delta_{j}+\alpha}\label{eq:distance}
\end{equation}

where $\delta_{k}\equiv\delta\left(v_{k}\right)$, $\delta_{j}\equiv\delta\left(v_{j}\right)$.
$\alpha$ is a parameter to balance the two distances. We utilize the
Kuhn--Munkres algorithm \cite{kuhn,munkres} to solve this perfect
maximum matching problem. Once getting the perfect matching, we averagely
merge the matched nodes and edges as shown in Figure \ref{fig:Matching-and-merging.}.
The newly generated edge is the average of original edges between
nodes involved by the new nodes. In this way, we can fuse room-wise
HOZ graphs two-by-two each time and finally get the compositive graph,
which is defined as scene-wise HOZ graph $G_{s}\left(V_{s},E_{s}\right)$.
Algorithm \ref{alg:Category-wise-HOZ-graph} summarizes the construction of scene-wise HOZ graph. All scene-wise HOZ graphs constitute
our final HOZ graph.

\subsection{HOZ Graph Updating and Embedding\label{subsec:HOZ-Graph-Embedding}}

\subsubsection{Zone Updating and Embedding}

With all training data, we can obtain a general HOZ graph $G\left(V,E\right)$
for the seen environments. Since different environments have various
layouts, especially in the new unseen environment, it is difficult
to construct  a precise graph from scratch. Therefore, we first learn
a general HOZ graph, and then propose an online-learning method to
update current zone node according to agent's real-time view. In this
way, the initial HOZ graph will evolve towards current environment. 
Note that the zone update only holds for an episode and each episode starts from the initial HOZ graph.

Through object detection, the agent obtains a bag-of-objects feature
$f_{t}\in\mathbb{R}^{N\times1}$ for object categories appearing in
the egocentric view $o_{t}$ at timestamp $t$. According to the visual
feature $f_{t}$, target object $p\in P$ and HOZ graph $G\left(V,E\right)$,
the agent calculates the current zone $Z_{c}$, target zone $Z_{t}$
and sub-goal zone $Z_{sub}$, which will be detailed in
Section \ref{subsec:Zone-Localization-and}. These zone indicator
vectors $Z_{c}, Z_{t}, Z_{sub}\in\mathbb{R}^{K\times1}$ are one-hot vectors
that only activate representative zones. The proposed HOZ graph $G\left(V,E\right)$
is embedded with GCN. At time $t=0$, the input matrix $\delta\left(V^{0}\right)\in\mathbb{R}^{K\times N}$
represents embedded features for all zone nodes $V$. Then $\delta\left(V^{t}\right)$
will be updated based on $f_{t}$, which can be formulated
as
\begin{equation}
\delta\left(V^{t}\right)=\lambda Z_{c}f_{t}^{T}+\left(I-\lambda Z_{c}Z_{c}^{T}\right)\delta\left(V^{t-1}\right)\label{eq:update}
\end{equation}

where $\lambda$ is a learnable parameter that determines the current
observation's impact on the general HOZ graph. Following \cite{Thomas_GCN},
we perform normalization on edges $E$ and obtain $\hat{E}$. With
updated zone nodes $\delta\left(V^{t}\right)$, adjacent relationship
$\hat{E}$, our GCN outputs a node-level representation $H_{z}\in\mathbb{R}^{K\times N}$
as the zones embedding
\begin{equation}
H_{z}=\sigma\left(\hat{E}\delta\left(V^{t}\right)W_{z}\right)
\end{equation}

where $\sigma\left(\cdot\right)$ denotes the ReLU activation function,
and $W_{z}\in\mathbb{R}^{N\times N}$ is the parameter of GCN layers.
Then we take the encoded vector $H_{z}^{T}Z_{sub}$ as the output
of zones layer, which informs agent about the next sub-goal zone and
its relative position to other zones.

\begin{figure*}
\begin{centering}
\includegraphics[scale=0.63]{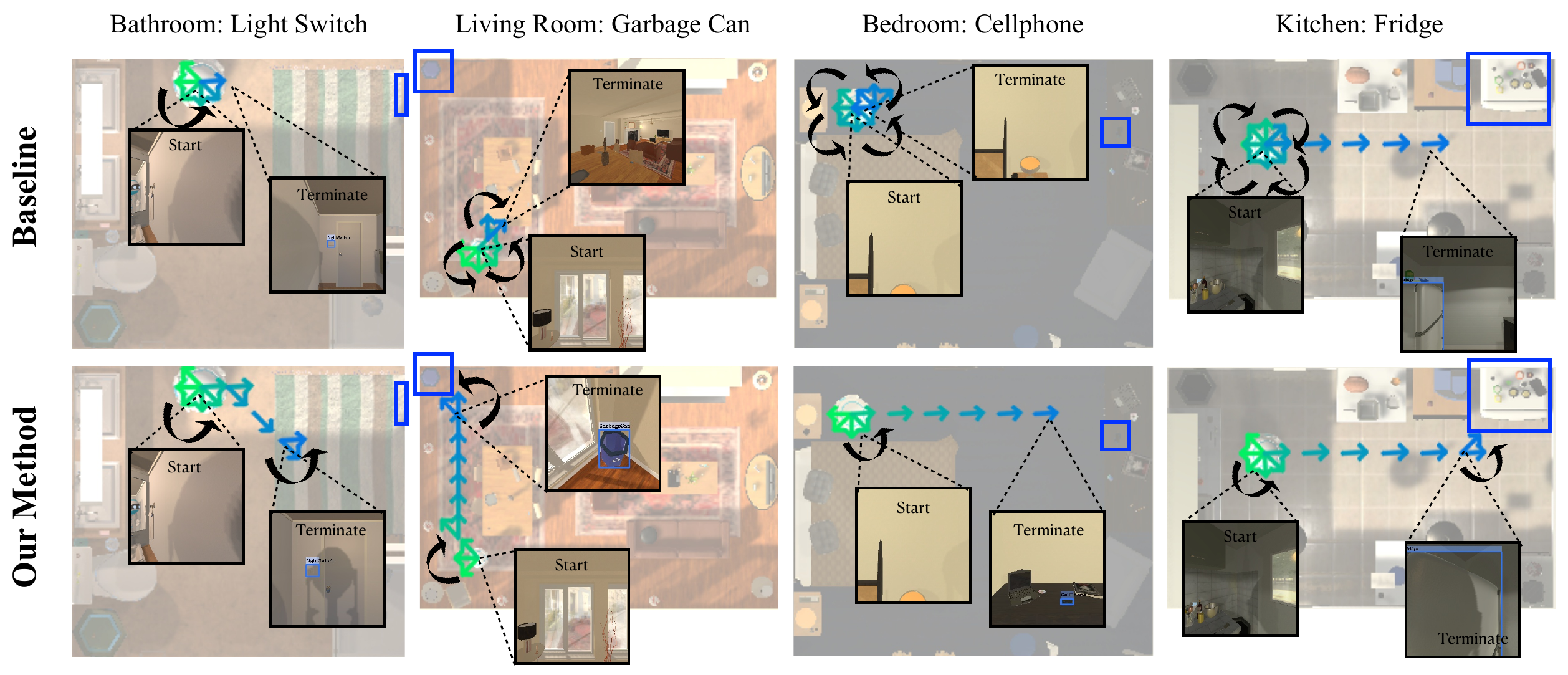}
\par\end{centering}
\caption{\label{fig:Visual}\textbf{Visualization in testing environments.}
Black arrows represent rotations. The trajectory of the agent is illustrated
with green and blue arrows, where green is the beginning and blue
is the end.}

\vspace{-10pt}
\end{figure*}

\subsubsection{Object Embedding}

Following \cite{ECCV_relation_graph}, we set up objects layer with
objects as nodes and relations between objects as edges, and
encode them with GCN. For current egocentric view, we can get the
detection feature $F_{t}=\left\{f_{t}^{b},f_{t}^{s},f_{t}^{v}\right\} $,
where $f_{t}^{b}\in\mathbb{R^{\mathrm{N\times4}}}$ is bounding box
position, $f_{t}^{s}\in\mathbb{R}^{N\times1}$ is confidence score
and $f_{t}^{v}\in\mathbb{R^{\mathrm{N\times512}}}$ is the visual
feature of objects. If multiple instances belonging to the same category
appear simultaneously, the one with the highest confidence score provided
by the detector will be selected. Define $X_{o}=\left[f_{t}^{b},f_{t}^{s},p\right]\in\mathbb{R}^{\mathrm{N\times6}}$
as the input of GCN, where $p\in\mathbb{R}^{N\times1}$ is a one-hot
vector representing the target object. The GCN outputs
\begin{equation}
H_{o}=\sigma\left(AX_{o}W_{o}\right)
\end{equation}

Both the adjacency matrix $A$ and the GCN network parameter $W_{o}\in\mathbb{R}^{6\times N}$
need to be learned. Then we integrate 
$H_{o}f_{t}^{v}$ as the objects embedding, which provides
object-level information.

\section{Navigation Policy}

\subsection{Zone Localization and Navigation Planning\label{subsec:Zone-Localization-and}}

\paragraph{Current zone}

We compare current view bag-of-objects vector $f_{t}$ with the nodes
in the pre-learned HOZ graph $G\left(V,E\right)$, and take the most
similar node as the current zone, which can be formulated as
\begin{equation}
Z_{c}=\chi^{K}\left(\underset{k}{argmin}\left(d\left(f_{t},\delta\left(v_{k}\right)\right)\right)\right),v_{k}\in V
\end{equation}

where $\chi^{K}\left(\cdot\right)$ is an indicator that
produces a one-hot vector $\chi^{K}\left(i\right)=\left[x_{1},\ldots x_{K}\right]^{T}$,
where $x_{i}=1,x_{j\neq i}=0$. $d\left(\cdot\right)$ is defined in
Equation \ref{eq:distance}. Then the HOZ graph is updated by the current zone $Z_{c}$
and the real-time feature $f_{t}$
(Equation \ref{eq:update}).

\paragraph{Target zone}

We take the node with the highest occurrence probability of the target
object as the target zone.
\begin{equation}
Z_{t}=\chi^{K}\left(\underset{k}{argmax}\left(\delta\left(v_{k}\right)^{T}p\right)\right),v_{k}\in V
\end{equation}

\paragraph{Sub-goal zone}

To navigate agent from current zone to target zone, we search for
a path with the maximum connection probability. If an edge has a higher
value, the two related zones are more likely to be adjacent so that
agent can easily arrive. Besides, when the target zone is far away
from the current zone or is not visible in the current view, the agent
may not be well guided. Therefore, we take the second child zone starting
from the current zone on this path as the sub-goal zone, which provides
information about where to go next. Our goal is to find an optimal
maximum connectivity path $\varGamma=\left\{ v_{\tau_{0}},v_{\tau_{1}},\ldots,v_{\tau_{T}}\right\} $,
where $\tau_{i}\in\left\{ 1,\ldots,K\right\} $ denotes the node index
and $v_{\tau_{0}}$ represents the current zone and $v_{\tau_{T}}$
represents the target zone, so that the connection probability along
the path is maximized as: 
\begin{equation}
\varGamma^{*}=\underset{\varGamma}{argmax}\mathop{\Pi_{i=1}^{T}e\left(v_{\tau_{i-1}},v_{\tau_{i}}\right)}
\end{equation}

After obtaining $\varGamma^{*}$, we can get the sub-goal zone $Z_{sub}=\chi^{K}\left(\tau_{1}^{*}\right)$.
Whenever the current zone changes, the network will adaptively replan
an optimal path and a sub-goal zone.

\begin{table*}
\setlength{\tabcolsep}{7pt} \renewcommand{\arraystretch}{1}\caption{\label{tab:Comparisons-of-embedding}\textbf{Comparisons with sub-goal
zone and target zone (\%).} The input of zone-to-action LSTM during training
and testing is set to the sub-goal zone (S) or target zone (T) respectively.}

\begin{centering}
\begin{tabular}{c|cc|cc|ccc|ccc}
\hline 
\multirow{2}{*}{Method} & \multicolumn{2}{c|}{Training} & \multicolumn{2}{c|}{Testing } & \multicolumn{3}{c|}{ALL} & \multicolumn{3}{c}{$L\geq5$}\tabularnewline
 & S & T & S & T & SR & SPL & SAE & SR & SPL & SAE\tabularnewline
\hline 
\hline 
Baseline &  &  &  &  & 57.35$_{\pm1.92}$ & 33.78$_{\pm1.33}$ & 19.02$_{\pm1.36}$ & 45.77$_{\pm2.17}$ & 30.65$_{\pm1.01}$ & 20.04$_{\pm1.87}$\tabularnewline
HOZ & $\surd$ &  & $\surd$ &  & \textbf{70.57}$_{\pm1.11}$ & \textbf{40.84$_{\pm1.12}$} & \textbf{27.19}$_{\pm1.96}$ & \textbf{61.52}$_{\pm1.47}$ & \textbf{40.46$_{\pm0.63}$} & \textbf{29.61}$_{\pm1.08}$\tabularnewline
\hline 
HOZ & $\surd$ &  &  & $\surd$ & 69.04$_{\pm1.07}$ & 40.07$_{\pm1.04}$ & 26.19$_{\pm0.95}$ & 59.27$_{\pm1.33}$ & 39.12$_{\pm0.83}$ & 28.34$_{\pm1.24}$\tabularnewline
HOZ &  & $\surd$ &  & $\surd$ & 69.16$_{\pm1.15}$ & 39.05$_{\pm0.88}$ & 26.04$_{\pm0.91}$ & 60.28$_{\pm1.42}$ & 38.61$_{\pm0.86}$ & 29.08$_{\pm0.98}$\tabularnewline
\hline 
\end{tabular}
\par\end{centering}
\vspace{-5pt}
\end{table*}

\subsection{Policy Learning\label{subsec:Policy-learning}}

\paragraph{Action policy}

The conventional works \cite{savn,ECCV_relation_graph,scene_priors,Target-driven_navigation}
learn a policy $\pi\left(a_{t}|o_{t},p\right)$ based on current observation.
While in our work, we learn a zone-to-action LSTM action policy $\pi_{z}\left(a_{t}|S_{t},p\right)$,
where $S_{t}$ is the joint representation of current observation
$o_{t}$, the sub-goal zone embedding $H_{z}^{T}Z_{sub}$ and object embedding $H_{o}f_{t}^{v}$. Following
\cite{Target-driven_navigation,Piotr_ICLR17} formulating
this task as a reinforcement learning problem, we optimize the LSTM
via the Asynchronous Advantage Actor-Critic (A3C) algorithm \cite{A3C}
that learns policy function and value function by minimizing navigation
loss $\mathcal{L}_{nav}$ to maximize the reward. The policy function
outputs $a_{t}$, representing actions probability at each time, and
the value function is used to train the policy network.

\paragraph{Done reminder}

To remind agent to stop in time when it encounters the target object,
we propose the done reminder. Combining objects detection confidence
$f_{t}^{s}$ and the target object $p$, we weight $a_{t}$ with $\beta p^{T}f_{t}^{s}$
to represent the effect of $done$ action ($\beta$ is a learnable
parameter). In this way, we can get the final action output $\hat{a}_{t}$.

\section{Experiments}

\begin{figure}
\begin{centering}
\includegraphics[scale=0.14]{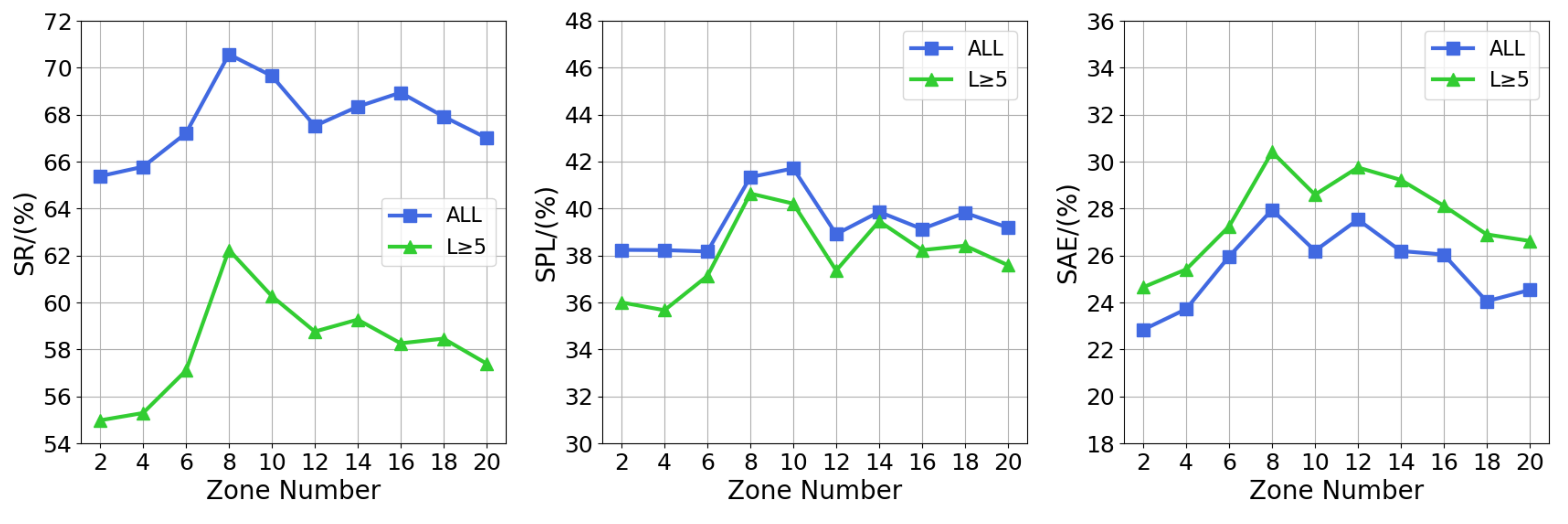}
\par\end{centering}
\caption{\label{fig:zone_number}\textbf{Ablation results of zone number. }We
evaluate the impact of the zone number (cluster number) on navigation
metrics such as SR, SPL, and SAE.}

\vspace{-10pt}
\end{figure}

\subsection{Experiment Setup}

We evaluate our methods on AI2-Thor simulator \cite{ai2thor}, which
provides near photo-realistic observation in 3D indoor scenes. AI2-Thor
contains a total of 120 scenes in 4 types: living room, kitchen, bedroom,
and bathroom, where spatial layout, object types and appearance are
all different. Following the setting in \cite{savn}, a subset of
22 types of objects is considered, ensuring that each scene contains
at least four objects. For each scene type, we choose 20 rooms for
training, 5 for validation, and 5 for test. 

\subsection{Implementation Details}

The baseline is the A3C \cite{A3C} navigation policy with
a simple visual embedding layer to encode inputs. We train our models
with $12$ asynchronous workers, in a total of 6M navigation episodes.
In policy learning, the agent receives a $-0.01$ penalty for each step and a reward of $5$ if the episode is successful. 
We use Adam optimizer \cite{Adam}
to update our network parameters with a learning rate of $10^{-4}$.
ResNet18 \cite{resnet} pretrained on ImageNet \cite{ImageNet} is
used as our backbone to extract the features of each egocentric view.
In the HOZ graph construction, we finetune Faster-RCNN \cite{Faster_R-CNN}
architecture on 50\% training data of AI2-Thor. The hyper-parameters
in our model are initialized to $\varepsilon=0.25$, $\alpha=0.1$
and $\beta=0.6$. 

For evaluation, we randomly select agent's initial starting position
and the target object, and repeatedly run 5 trials. We report results (with average and variance) for all targets (All) and a subset
of targets ($L\geq5$) whose optimal trajectory length is longer than
5.

\subsection{Evaluation Metrics}

We use Success Rate (SR), Success Weighted by Path Length (SPL) \cite{eval_embodied},
and Success Weighted by Action efficiency (SAE) metrics to evaluate
our model. SR refers to the success rate of agent in finding the target
object, which is formulated as $SR=\frac{1}{N}\sum_{n=1}^{N}Suc_{n}$,
where $N$ is the total number of episodes and $Suc_{n}$ is an indicator
function to indicate whether the $n$-th episode succeeds. SPL considers
both the success rate and the path length. It is defined as $SPL=\frac{1}{N}\sum_{i=1}^{N}Suc_{i}\frac{L_{i}^{*}}{max\left(L_{i},L_{i}^{*}\right)}$,
where $L_{i}$ is the actual path length and $L_{i}^{*}$ represents
the shortest path provided by the simulator. Although SPL calculates
the proximity between the path and the optimal path, it ignores the
efficiency of action sequence. For instance, unnecessary rotations
take time and reduce efficiency, which are not considered in SPL.
Therefore, we propose the SAE metric to measure the efficiency of
all actions. It is formulated as $SAE=\frac{1}{N}\sum_{i=1}^{N}Suc_{i}\frac{\sum_{t=0}^{T}\mathbb{I}\left(a_{t}^{i}\in\mathcal{A}_{change}\right)}{\sum_{t=0}^{T}\mathbb{I}\left(a_{t}^{i}\in\mathcal{A}_{all}\right)}$,
where $\mathbb{I}\left(\cdot\right)$ is the indicator function, $a_{t}^{i}$
is agent's action at time $t$ in episode $i$, $\mathcal{A}_{all}$
is the set of all action categories and $\mathcal{A}_{change}$ refers
to those actions that can change agent's location. In our settings
$\mathcal{A}_{change}=\left\{ MoveAhead\right\} $.

\subsection{Ablation Study}

\begin{table*}
\setlength{\tabcolsep}{4pt} \renewcommand{\arraystretch}{1}

\caption{\label{tab:ablation}\textbf{The ablation study of different components (\%).}
We evaluate the effect of various modules.
These modules include the scene layer (Scene), the zone layer (Zone),
the object layer (Object) in Section \ref{subsec:HOZ-Graph-Embedding} and the done reminder (Reminder) in Section \ref{subsec:Policy-learning}.}

\noindent \begin{centering}
\begin{tabular}{ccccc|ccc|ccc}
\hline 
\multirow{2}{*}{Baseline} & \multirow{2}{*}{Scene} & \multirow{2}{*}{Zone} & \multirow{2}{*}{Object} & \multirow{2}{*}{Reminder} & \multicolumn{3}{c|}{All} & \multicolumn{3}{c}{$L\geq5$}\tabularnewline
 &  &  &  &  & SR & SPL & SAE & SR & SPL & SAE\tabularnewline
\hline 
\hline 
$\surd$ &  &  &  &  & 57.35$_{\pm1.92}$ & 33.78$_{\pm1.33}$ & 19.02$_{\pm1.36}$ & 45.77$_{\pm2.17}$ & 30.65$_{\pm1.01}$ & 20.04$_{\pm1.87}$\tabularnewline
$\surd$ &  &  & $\surd$ &  & 65.12$_{\pm1.03}$ & 37.86$_{\pm0.93}$ & 24.36$_{\pm0.91}$ & 53.42$_{\pm1.43}$ & 35.37$_{\pm0.71}$ & 25.32$_{\pm1.04}$\tabularnewline
$\surd$ & $\surd$ &  & $\surd$ &  & 65.81$_{\pm1.11}$ & 38.83$_{\pm0.59}$ & 22.45$_{\pm0.99}$ & 57.23$_{\pm0.93}$ & 36.25$_{\pm0.65}$ & 25.53$_{\pm0.87}$\tabularnewline
$\surd$ &  &  & $\surd$ & $\surd$ & 66.73$_{\pm1.01}$ & 37.82$_{\pm0.83}$ & 24.81$_{\pm0.84}$ & 57.55$_{\pm1.19}$ & 36.48$_{\pm0.52}$ & 27.79$_{\pm1.07}$\tabularnewline
$\surd$ & $\surd$ & $\surd$ & $\surd$ &  & 70.57$_{\pm1.11}$ & \textbf{40.84$_{\pm1.12}$} & 27.19$_{\pm1.96}$ & 61.52$_{\pm1.47}$ & \textbf{40.46$_{\pm0.63}$} & 29.61$_{\pm1.08}$\tabularnewline
$\surd$ & $\surd$ & $\surd$ & $\surd$ & $\surd$ & \textbf{70.62$_{\pm1.70}$} & 40.02$_{\pm1.25}$ & \textbf{27.97$_{\pm2.01}$} & \textbf{62.75$_{\pm1.73}$} & 39.24$_{\pm0.56}$ & \textbf{30.14$_{\pm1.34}$}\tabularnewline
\hline 
\end{tabular}
\par\end{centering}
\vspace{-5pt}
\end{table*}

\begin{table}
\setlength{\tabcolsep}{2pt} \renewcommand{\arraystretch}{1}\caption{\label{tab:Comparisons-with-non-adaptive}\textbf{Comparisons with
the related works (\%).} Constrained by space, variance is
detailed in supplementary materials. }

\noindent \begin{centering}
\begin{tabular}{c|ccc|ccc}
\hline 
\multirow{2}{*}{Method} & \multicolumn{3}{c|}{All} & \multicolumn{3}{c}{$L\geq5$}\tabularnewline
 & SR & SPL & SAE & SR & SPL & SAE\tabularnewline
\hline 
\hline 
\multicolumn{7}{c}{Non-adaptive method}\tabularnewline
\hline 
\hline 
Random & 3.56 & 1.73 & 0.41 & 0.27 & 0.07 & 0.06\tabularnewline
A3C (baseline) & 57.35 & 33.78 & 19.02 & 45.77 & 30.65 & 20.04\tabularnewline
SP \cite{scene_priors} & 62.16 & 37.01 & 23.39 & 50.86 & 34.17 & 24.35\tabularnewline
ORG \cite{ECCV_relation_graph} & 66.38 & 38.42 & 25.36 & 55.55 & 36.26 & 27.53\tabularnewline
\hline 
\textbf{Ours (HOZ)} & \textbf{70.62} & \textbf{40.02} & \textbf{27.97} & \textbf{62.75} & \textbf{39.24} & \textbf{30.14}\tabularnewline
\hline 
\hline 
\multicolumn{7}{c}{Self-supervised method}\tabularnewline
\hline 
\hline 
SAVN \cite{savn} & 63.32 & 37.62 & 21.97 & 52.38 & 35.31 & 24.64\tabularnewline
ORG-TPN \cite{ECCV_relation_graph} & 67.31 & \textbf{39.53} & 23.07 & 57.41 & 38.27 & 26.37\tabularnewline
\hline 
\textbf{Ours (HOZ-TPN)} & \textbf{73.15} & 39.22 & \textbf{29.49} & \textbf{64.58} & \textbf{39.80} & \textbf{30.92}\tabularnewline
\hline 
\end{tabular}
\par\end{centering}
\vspace{-5pt}
\end{table}

\paragraph{Effectiveness of sub-goal zones}

As discussed in Section \ref{subsec:Zone-Localization-and}, besides
the target zone, we also consider the sub-goal zone. The ablation
study respectively trains the policy network with the sub-goal zone
and the target zone as illustrated in Table \ref{tab:Comparisons-of-embedding}
\textit{line2} and \textit{line}4. Compared to the target zone, sub-goal
zone can better guide agent efficiently. Training with the embedding
of sub-goal zone outperforms target zone by 1.41/1.24, 1.79/1.85 and 1.15/0.53 in SR, SPL and SAE (ALL/$L\geq5$, \%) respectively.

\paragraph{Impacts of the number of zones}

The cluster number is a hyper-parameter that specifies the zone number
in a scene. Figure \ref{fig:zone_number} indicates that performance
is reduced when the number of zones is either too large or too small.
Besides, a large zone number requires significant computing
resources when planning the path. The results suggest that the optimal
number of zones is 8. Therefore, the number of zones 
is set to 8 in the remaining evaluations.

\paragraph{Other ablation studies}

We dissect the proposed HOZ graph into different components. The ablation study in Table \ref{tab:ablation} demonstrates the efficacy of each component of our method. Specifically, it is observed that the object layer significantly improves the baseline performance. Additionally, scene and zone layers can considerably increase the performance on SPL and SAE metrics. Although the done reminder decreases the SPL metric, it increases the SR and SAE metrics, indicating that adding the done reminder lengthens the episodes. 
Overall, our method outperforms the baseline model with the gains of 13.27/16.98, 6.24/8.59 and 8.95/10.10  in SR, SPL and SAE (ALL/$L\geq5$, \%). The experimental results indicate that our method is capable of effectively guiding navigation in the unseen environments. 

In addition, considering that the construction of the scene-wise HOZ graph may be inconsistent due to different merging order of room-wise HOZ graph.
We test 20 different merging orders to get the variance of
0.83/0.81, 0.78/0.81, 0.81/0.82 in SR, SPL and SAE (ALL/$L\geq5$, \%). These results indicate that the merging-related potential
inconsistency has little effect on the navigation performance.

\subsection{Comparisons to the State-of-the-art}

Related works can be categorized into non-adaptive models \cite{ECCV_relation_graph,scene_priors}
and self-supervised models \cite{savn,ECCV_relation_graph}. Compared
with the non-adaptive methods in Table \ref{tab:Comparisons-with-non-adaptive},
our method outperforms the state-of-the-art by a large margin in
SR, SPL and SAE metrics. Particularly, we obtain the gains of 4.24/7.20, 1.60/2.98, 2.61/2.61 in SR, SPL and SAE (ALL/$L\geq5$, \%)
over the state-of-the-art model \cite{ECCV_relation_graph}.

Compared to the non-adaptive models, the self-supervised models are
updated with self-supervision in test. This self-supervision can somehow improve performance, but also consume additional computing resources. We also implement our methods with self-supervision (denoted as HOZ-TPN). 
In comparison to HOZ, HOZ-TPN improves SR but achieves equivalent results in SPL and SAE, which are more indicative of navigation efficiency.
The comparison between HOZ and HOZ-TPN (as well as ORG and ORG-TPN) demonstrates that while self-supervision may aid in successfully navigating to target objects, it also introduces additional actions.
More experimental results are detailed in supplementary materials. 

\paragraph{Case study}

Figure \ref{fig:Visual} qualitatively compares our HOZ with the baseline
model. In these scenarios, the agent is placed at an initial position
where the target object cannot be seen. The baseline model often falls
into rotations when the target object is not in the view. However,
our HOZ method helps the agent locate the current zone and offers
guidance from the current zone to the target zone, thus the agent
has better performance. Notably, with the guidance of sub-goal zone
, the agent equipped with our HOZ graph can choose a better rotation
direction than the baseline method.

\section{Conclusions}

We propose the hierarchical
object-to-zone (HOZ) graph that captures the prior knowledge of objects
in typical zones. The agent equipped with HOZ is capable of updating
prior knowledge, locating the target zone and planning the zone-to-zone
path. We also propose a new evaluation metric named Success weighted
by Action Efficiency (SAE) that measures the efficiency of actions.
Experimental results show that our approach outperforms baseline by
a large margin in SR, SPL and SAE metrics.

\paragraph*{{\small{}Acknowledgements: }}

\noindent {\small{}This work was supported by National Key Research and Development Project of New Generation Artificial Intelligence of China, under Grant 2018AAA0102500, in part by the National Natural Science Foundation of China under Grant 62032022, 61902378 and U1936203, in part by Beijing Natural Science Foundation under Grant L182054 and Z190020, in part by the Lenovo Outstanding Young Scientists Program, in part by the National Postdoctoral Program for Innovative Talents under Grant BX201700255.}{\small\par}

{\small
\bibliographystyle{ieee_fullname}
\bibliography{egbib}
}

\appendix

\section{Video Demo}

A video demo that visualizes the construction of HOZ graph, navigation
with HOZ graph and more case studies can be found at the following
url:

\url{https://drive.google.com/file/d/1UtTcFRhFZLkqgalKom6_9GpQmsJfXAZC/view?usp=sharing} 

\section{Navigation Target}

The target objects of different scenes in AI2THOR \cite{ai2thor}
are shown in Table \ref{tab:split_ai2thor}. Our training and testing
share the consistent target objects categories, though the testing
environments are new and unseen.

Considering that each environment in AI2THOR usually contains one
room, the agent navigation may be limited to short trajectories. Thus,
for longer trajectories object navigation, we also conduct experiments
on a more complex simulator RoboTHOR \cite{RoboTHOR}, which has 2.4
times larger area and 5.5 times longer trajectory length than AI2THOR.
The environment in RoboTHOR usually contains a variety of rooms. To
highlight the differences between AI2THOR and RoboTHOR, we define
each environment in AI2THOR as \textit{room} and that in RoboTHOR
as \textit{apartment.} In RoboTHOR, 12 objects categories are selected
as target objects for training and testing, involving \textit{Book, Bowl,
Chair, Plate, Television, Floor Lamp, Garbage Can, Alarm Clock, Desk
Lamp, Laptop, Pot, CellPhone}.  The experimental results are shown
in Section \ref{subsec:results_RoboTHOR}.

\begin{table}[t]
\setlength{\tabcolsep}{12pt}
\noindent \begin{centering}
\caption{\label{tab:split_ai2thor}\textbf{Object categories for navigation.}
The target objects categories of different room types in AI2THOR \cite{ai2thor}.}
\par\end{centering}
\noindent \centering{}%
\begin{tabular}{cc}
\hline 
Scenes & Objects\tabularnewline
\hline 
\hline 
\multirow{5}{*}{Kitchen} & Fridge, Light Switch, Pot,\tabularnewline
 & Coffee Machine, Sink, Pan,\tabularnewline
 & Chair, Plate, Bowl, Toaster,\tabularnewline
 & Stove Burner, Kettle,\tabularnewline
 & Microwave, Garbage Can\tabularnewline
\hline 
\multirow{4}{*}{Living Room} & FloorLamp, Chair, Plate,\tabularnewline
 & Light Switch, Garbage Can,\tabularnewline
 & Laptop, Remote Control, Book,\tabularnewline
 & Television, Desk Lamp\tabularnewline
\hline 
\multirow{3}{*}{Bedroom} & Book, Light Switch, Bowl,\tabularnewline
 & Desk Lamp, Laptop, Chair,\tabularnewline
 & Alarm Clock, Garbage Can,\tabularnewline
\hline 
\multirow{2}{*}{Bathroom} & Light Switch, Garbage Can,\tabularnewline
 & Sink\tabularnewline
\hline 
\end{tabular}
\vspace{-5pt}
\end{table}

\section{More Ablation Studies}

\begin{table*}
\setlength{\tabcolsep}{10pt}\caption{\label{tab:zone_node}\textbf{Comparisons with different information
used for clustering (\%).} The zone clustering is based on different information, including visual information $f$ (Visual) and 
location information $l$ (Location).}
\centering{}%
\begin{tabular}{cc|ccc|ccc}
\hline 
\multirow{2}{*}{Visual} & \multirow{2}{*}{Location} & \multicolumn{3}{c|}{ALL} & \multicolumn{3}{c}{$L\geq5$}\tabularnewline
 &  & SR & SPL & SAE & SR & SPL & SAE\tabularnewline
\hline 
\hline 
$\surd$ &  & \textbf{70.62$_{\pm1.70}$} & 40.02$_{\pm1.25}$ & \textbf{27.97$_{\pm2.01}$} & \textbf{62.75$_{\pm1.73}$} & \textbf{39.24}$_{\pm0.56}$ & \textbf{30.14$_{\pm1.34}$}\tabularnewline
$\surd$ & $\surd$ & 68.2\textbf{2$_{\pm1.54}$} & \textbf{40.48}$_{\pm1.07}$ & 25.81\textbf{$_{\pm1.78}$} & 60.63$_{\pm1.46}$ & 37.92$_{\pm0.48}$ & 29.09$_{\pm1.01}$\tabularnewline
\hline 
\end{tabular}
\end{table*}

\subsection{Clustering information}

\begin{table*}
\setlength{\tabcolsep}{12pt}\caption{\label{tab:Comparisons-of-embedding}\textbf{Comparisons with different
detection modules (\%).} We compare the impact of utilizing a
pre-trained detection model (Detection Pre) or the ground truth of
object detection (Detection GT).}

\centering{}%
\begin{tabular}{c|ccc|ccc}
\hline 
\multirow{2}{*}{Module} & \multicolumn{3}{c|}{ALL} & \multicolumn{3}{c}{$L\geq5$}\tabularnewline
 & SR & SPL & SAE & SR & SPL & SAE\tabularnewline
\hline 
\hline 
Detection Pre & 65.12$_{\pm1.03}$ & 37.86$_{\pm0.93}$ & 24.36$_{\pm0.91}$ & 53.42$_{\pm1.43}$ & 35.37$_{\pm0.71}$ & 25.32$_{\pm1.04}$\tabularnewline
Detection GT & \textbf{66.78$_{\pm0.73}$} & \textbf{55.91$_{\pm0.46}$} & \textbf{26.73$_{\pm0.26}$} & \textbf{55.02$_{\pm0.68}$} & \textbf{48.73$_{\pm0.31}$} & \textbf{30.23$_{\pm0.33}$}\tabularnewline
\hline 
\end{tabular}
\vspace{-5pt}
\end{table*}

In our method, we sample a set of features $(f,l)$ according to the
observations in the environments, where $f$ is a bag-of-objects vector
representing objects categories detected in view, and $l$ represents
the sample location. Then we implement feature clustering on $f$,
and each obtained cluster serves as a zone node in room-wise HOZ.
That is to say, our zone node is only based on visual information.
In order to further explore the impact of clustering, we introduce
the additional location information and cluster on both $(f,l)$.
Table \ref{tab:zone_node} demonstrates the navigation performance
with these two clustering methods. The results show that clustering
on both visual and location information drops 2.40/2.12\% and 2.16/1.05\% in SR and SAE and slightly improves in SPL, suggesting that
the additional location information narrows the range of our proposed
\emph{zone}. In other words, our HOZ (clustering on visual information)
treats all regions where agent can observe similar objects with a
specified direction as a zone, while clustering with both visual and
location information restrains the zone region merely around these
objects. Thus, location is more like a constraint rather than helpful
information, limiting the visual generalization of the proposed HOZ graph.
When the target object is not in view, agent needs to search more
zones until discovering the target. It is obviously inefficient so
that we obtain zone nodes for HOZ only based on visual information.

\subsection{Object detection module}

Table \ref{tab:Comparisons-of-embedding} shows the impact of different
detection modules on navigation performance, where \textit{Detection
Pre } indicates that the detection module is pre-trained with labeled
egocentric images sampled in simulator, and \textit{Detection GT}
 indicates that the detection module is ground truth provided by simulator.
The ablation with ground truth detection improves performance by 1.66/1.60 , 2.37/4.91 and \textbf{18.05}/\textbf{13.36} in
SR, SAE and SPL (ALL/$L\geq5$, \%) respectively.  The results demonstrate
that accurately recognizing more objects can help agent navigate successfully
in shorter trajectories. It is easy to understand because agent can
take the most likely action at each step to obtain the high SPL. However,
since the navigation task includes multiple decision steps, its success
rate does not rely on taking the perfect action at each step. As long
as most actions are reasonable, the agent can still achieve success.
So the approximate results on SR and SAE indicate that our HOZ graph still makes
sense in guiding unseen object navigation.

\subsection{The ablations of graph settings}

Since our HOZ graph adds more parameters to the model, we perform
additional ablations of zone nodes and edges, as indicated in Table
\ref{tab:More-ablations}. To assess if the gain in network performance
is due to the increased number of parameters or the information contained
in the HOZ graph's nodes and edges, We respectively set the edges and nodes
of the HOZ graph to random. The experimental results show
that the control experiments with random settings perform worse than
the original value, demonstrating the efficacy of zone information
(nodes) and spatial priors (edges).

\begin{figure}
\begin{centering}
\includegraphics[scale=0.25]{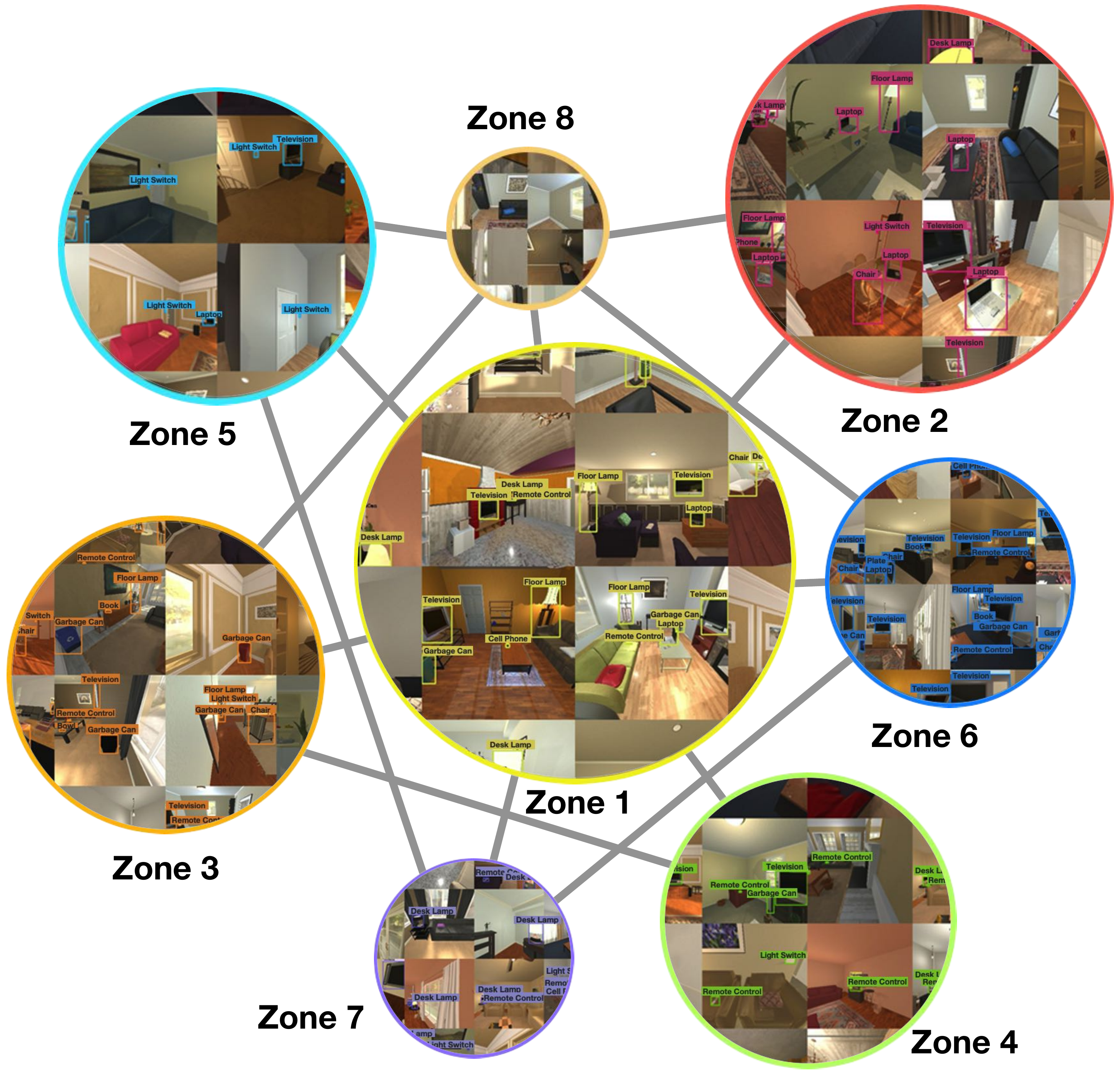}
\par\end{centering}
\caption{\label{fig:Scene-Zone-Graph}\textbf{Zones nodes of Hierarchical Object-to-Zone
Graph.} 8 different colors represent different zones. To highlight
the objects contained in these zones, we mark them with bounding boxes.}
\vspace{-10pt}
\end{figure}

\section{More comparisons with the related works}
\begin{table*}[t]
\setlength{\tabcolsep}{8pt}

\caption{\label{tab:More-ablations}\textbf{More ablations of graph settings
(\%).} The parameters of nodes or edges are randomly set (R) or kept
(K).}

\centering{}%
\begin{tabular}{cc|ccc|ccc}
\hline 
\multirow{2}{*}{Nodes} & \multirow{2}{*}{Edges} & \multicolumn{3}{c|}{ALL} & \multicolumn{3}{c}{$L\geq5$}\tabularnewline
 &  & SR & SPL & SAE & SR & SPL & SAE\tabularnewline
\hline 
\hline 
\multirow{2}{*}{R} & R & $67.81_{\pm0.62}$ & $38.92_{\pm0.22}$ & $24.13_{\pm0.35}$ & $57.84_{\pm0.81}$ & $38.22_{\pm0.44}$ & $24.02_{\pm0.52}$\tabularnewline
 & K & $68.52_{\pm1.05}$ & $39.83_{\pm0.52}$ & $26.52_{\pm0.62}$ & $58.61_{\pm0.82}$ & $38.73_{\pm0.62}$ & $28.73_{\pm0.53}$\tabularnewline
\hline 
\multirow{2}{*}{K} & R & $69.33_{\pm0.32}$ & $39.71_{\pm0.32}$ & $26.63_{\pm0.13}$ & $59.93_{\pm0.53}$ & $39.14_{\pm0.45}$ & $29.01_{\pm0.312}$\tabularnewline
 & K & $\textbf{70.47}_{\pm0.35}$ & $\textbf{40.66}_{\pm0.47}$ & $\textbf{27.85}_{\pm0.44}$ & $\textbf{62.17}_{\pm0.26}$ & $\textbf{40.14}_{\pm0.46}$ & $\textbf{30.33}_{\pm0.25}$\tabularnewline
\hline 
\end{tabular}
\end{table*}

\begin{figure*}
\noindent \begin{centering}
\includegraphics[angle=90,scale=0.65]{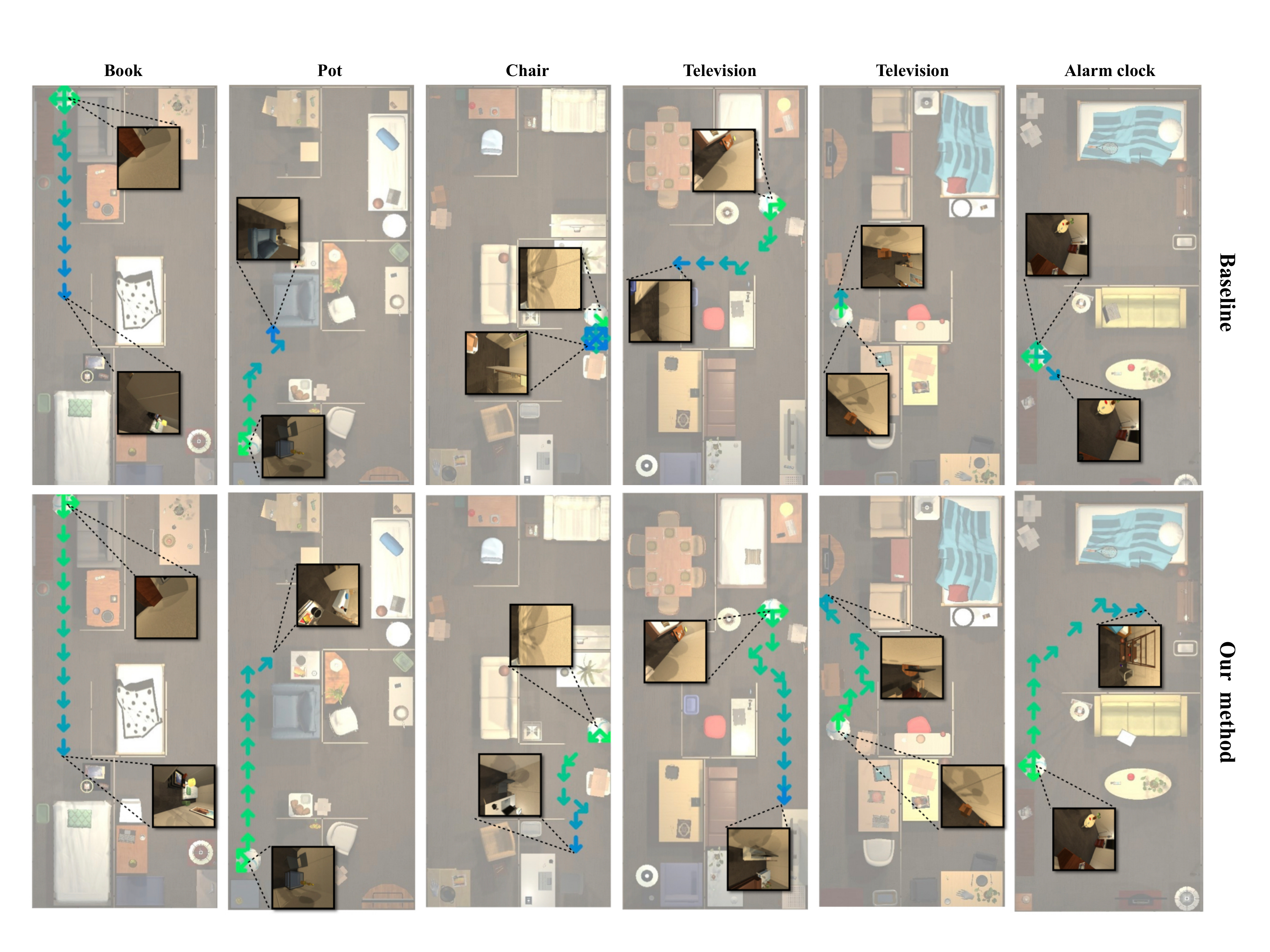}
\par\end{centering}
\caption{\label{fig:Visualization}\textbf{Visualization of trajectory in RoboTHOR.}
Black arrows represent rotations. The trajectory of the agent is illustrated
with green and blue arrows, where green is the beginning and blue
is the end.}
\end{figure*}

\begin{table*}[t]
\setlength{\tabcolsep}{11pt} \renewcommand{\arraystretch}{1}

\caption{\label{tab:Comparisons-with-non-adaptive}\textbf{Comparisons with
the related works in RoboTHOR \cite{RoboTHOR} (\%).} We repeat
the evaluations similar to AI2-Thor on RoboTHOR.}

\noindent \centering{}%
\begin{tabular}{c|ccc|ccc}
\hline 
\multirow{2}{*}{Method} & \multicolumn{3}{c|}{ALL} & \multicolumn{3}{c}{$L\geq5$}\tabularnewline
 & SR & SPL & SAE & SR & SPL & SAE\tabularnewline
\hline 
\hline 
\multicolumn{7}{c}{Non-adaptive method}\tabularnewline
\hline 
\hline 
Random & 0.00$_{\pm0.00}$ & 0.00$_{\pm0.00}$ & 0.00$_{\pm0.00}$ & 0.00$_{\pm0.00}$ & 0.00$_{\pm0.00}$ & 0.00$_{\pm0.00}$\tabularnewline
A3C (baseline) & 26.41$_{\pm0.52}$ & 16.61$_{\pm0.34}$ & 13.15$_{\pm0.43}$ & 17.42$_{\pm0.21}$ & 12.23$_{\pm0.66}$ & 10.94$_{\pm0.35}$\tabularnewline
SP \cite{scene_priors} & 28.04$_{\pm0.33}$ & 17.63$_{\pm0.26}$ & 14.23$_{\pm0.25}$ & 21.66$_{\pm0.32}$ & 15.14$_{\pm0.46}$ & 13.27$_{\pm0.34}$\tabularnewline
ORG \cite{ECCV_relation_graph} & 29.61$_{\pm0.71}$ & 19.23$_{\pm0.94}$ & 14.72$_{\pm0.64}$ & 22.53$_{\pm0.55}$ & 15.73$_{\pm0.86}$ & 13.82$_{\pm0.44}$\tabularnewline
\hline 
\textbf{Ours (HOZ)} & \textbf{32.27$_{\pm1.14}$} & \textbf{20.48$_{\pm0.63}$} & \textbf{17.18$_{\pm0.42}$} & \textbf{24.83$_{\pm0.72}$} & \textbf{16.89$_{\pm0.50}$} & \textbf{15.62$_{\pm0.55}$}\tabularnewline
\hline 
\hline 
\multicolumn{7}{c}{Self-supervised method}\tabularnewline
\hline 
\hline 
SAVN \cite{savn} & 28.42$_{\pm0.41}$ & 17.82$_{\pm0.33}$ & 13.91$_{\pm0.24}$ & 22.13$_{\pm0.32}$ & 15.34$_{\pm0.45}$ & 13.01$_{\pm0.24}$\tabularnewline
ORG-TPN \cite{ECCV_relation_graph} & 30.01$_{\pm1.22}$ & 20.51$_{\pm0.74}$ & 14.52$_{\pm0.93}$ & 22.25$_{\pm0.63}$ & 16.64$_{\pm0.35}$ & 13.83$_{\pm0.45}$\tabularnewline
\hline 
\textbf{Ours (HOZ-TPN)} & \textbf{33.28$_{\pm1.62}$} & \textbf{22.13$_{\pm0.91}$} & \textbf{16.66$_{\pm0.62}$} & \textbf{24.98$_{\pm1.32}$} & \textbf{18.05$_{\pm0.64}$} & \textbf{15.57$_{\pm0.76}$}\tabularnewline
\hline 
\end{tabular}
\end{table*}

\begin{table*}
\setlength{\tabcolsep}{11pt}\caption{\label{tab:AI2Thor}\textbf{Comparisons with the related works in AI2THOR (\%).}
These results are the supplement for Table 3 in the main text.}

\noindent \centering{}%
\begin{tabular}{c|ccc|ccc}
\hline 
\multirow{2}{*}{Method} & \multicolumn{3}{c|}{All} & \multicolumn{3}{c}{$L\geq5$}\tabularnewline
 & Suc. & SPL & SAE & Suc. & SPL & SAE\tabularnewline
\hline 
\hline 
\multicolumn{7}{c}{Non-adaptive method}\tabularnewline
\hline 
\hline 
Random & 3.56$_{\pm2.74}$ & 1.73$_{\pm1.52}$ & 0.41$_{\pm0.52}$ & 0.27$_{\pm0.22}$ & 0.07$_{\pm0.06}$ & 0.06$_{\pm0.05}$\tabularnewline
A3C (baseline) & 57.35$_{\pm1.92}$ & 33.78$_{\pm1.33}$ & 19.02$_{\pm1.36}$ & 45.77$_{\pm2.17}$ & 30.65$_{\pm1.01}$ & 20.04$_{\pm1.87}$\tabularnewline
SP \cite{scene_priors} & 62.16$_{\pm0.70}$ & 37.01$_{\pm0.68}$ & 23.39$_{\pm0.69}$ & 50.86$_{\pm0.34}$ & 34.17$_{\pm0.85}$ & 24.35$_{\pm0.74}$\tabularnewline
ORG \cite{ECCV_relation_graph} & 66.38$_{\pm0.95}$ & 38.42$_{\pm0.22}$ & 25.36$_{\pm0.43}$ & 55.55$_{\pm1.89}$ & 36.26$_{\pm0.39}$ & 27.53$_{\pm0.48}$\tabularnewline
\hline 
\textbf{Ours (HOZ)} & \textbf{70.62$_{\pm1.70}$} & \textbf{40.02$_{\pm1.25}$} & \textbf{27.97$_{\pm2.01}$} & \textbf{62.75$_{\pm1.73}$} & \textbf{39.24$_{\pm0.56}$} & \textbf{30.14$_{\pm1.34}$}\tabularnewline
\hline 
\hline 
\multicolumn{7}{c}{Self-supervised method}\tabularnewline
\hline 
\hline 
SAVN \cite{savn} & 63.32$_{\pm1.17}$ & 37.62$_{\pm0.86}$ & 21.97$_{\pm0.21}$ & 52.38$_{\pm0.73}$ & 35.31$_{\pm0.79}$ & 24.64$_{\pm0.52}$\tabularnewline
ORG-TPN \cite{ECCV_relation_graph} & 67.31$_{_{\pm1.14}}$ & \textbf{39.53$_{\pm1.01}$} & 23.07$_{\pm0.24}$ & 57.41$_{\pm0.71}$ & 38.27$_{\pm0.63}$ & 26.37$_{\pm0.57}$\tabularnewline
\hline 
\textbf{Ours (HOZ-TPN)} & \textbf{73.15$_{\pm1.01}$} & 39.22$_{\pm1.27}$ & \textbf{29.49$_{\pm0.11}$} & \textbf{64.58$_{\pm0.74}$} & \textbf{39.80$_{\pm0.57}$} & \textbf{30.92$_{\pm0.40}$}\tabularnewline
\hline 
\end{tabular}
\end{table*}

\subsection{Experiments on RoboTHOR \label{subsec:results_RoboTHOR}}

For longer trajectories object navigation, we also conduct experiments
on RoboTHOR \cite{RoboTHOR} simulator.\textit{ }RoboTHOR consists
of 89 apartments, 75 for training and validation, while the testing
data have not yet been made public. Therefore, we choose 60 apartments
for training, 5 for validation and 10 for testing. Since the regions
in RoboTHOR are simply separated with several clapboard, we treat
each apartment as a whole rather than subdividing it into scattered
scenes. Therefore, different from the construction of scene-wise HOZ
graph in AI2THOR, we build apartment-wise HOZ graph in RoboTHOR and
establish a unified HOZ graph combing all apartments. 

Table \ref{tab:Comparisons-with-non-adaptive} illustrates that our
method still outperforms the state-of-the-art with a large margin
by 2.66/2.30 in SR, 1.25/1.16 in SPL and 2.46/1.80 in SAE
metric (ALL/$L\geq5$, \%). Besides, compared with self-supervised methods,
our method equipped with the equal self-supervised adaptive module
also gains significant improvement of 3.27/2.73 in SR, 1.62/1.41 in SPL and 2.14/1.74 in SAE metric (ALL/$L\geq5$, \%).

In addition, we supplement the experimental results of variance for
Table 3 in the main text. The complete experimental results are shown in Table \ref{tab:AI2Thor}.

\subsection{Comparisons with semantic map}

In addition, Chaplot et al. \cite{Chaplot_NIPS20} attempt to construct
the episodic semantic map and use it to explore the unseen environment.
Different from our method that only relies on RGB input, the semantic
map is constructed based on a variety of inputs, including RGB-D input,
segmentation mask and GPS coordinate. We evaluate the HOZ graph and
the semantic map in Gibson \cite{Gibson}, where all methods utilize the RGB-D
input, segmentation mask and GPS coordinate. As indicated in Table \ref{tab:Gibson},
since the SLAM-based method processes multiple inputs more completely,
the performance of the baseline with the HOZ graph is slightly inferior
than SemExp. However, incorporating the HOZ graph for SemExp improves the SR,
SPL and SAE by 1.18, 0.34, 0.41 (ALL, \%) respectively, indicating that
the HOZ graph and SLAM-based method learn complementary information.
The experimental results demonstrate that the HOZ graph is also effective
when combined with SLAM-based methods. 

\begin{table}
\setlength{\tabcolsep}{5pt}

\caption{\label{tab:Gibson}\textbf{Comparisons with the semantic map in Gibson
(\%). }The baseline is the A3C model with a simple visual embedding
layer to encode various inputs. Since the path lengths of all episodes are
larger than 5, the subset of $L\protect\geq5$ is excluded.}

\centering{}%
\begin{tabular}{c|ccc}
\hline 
Method & SR & SPL & SAE\tabularnewline
\hline 
\hline 
Baseline + HOZ & 43.47$_{\pm0.51}$ & 12.88$_{\pm0.36}$ & 11.67$_{\pm0.51}$\tabularnewline
\hline 
SemExp \cite{Chaplot_NIPS20} & 44.01$_{\pm0.47}$ & 14.34$_{\pm0.42}$ & 12.32$_{\pm0.43}$\tabularnewline
\hline 
\textbf{SemExp + HOZ} & \textbf{45.19}$_{\pm0.35}$ & \textbf{14.68}$_{\pm0.38}$ & \textbf{12.73}$_{\pm0.45}$\tabularnewline
\hline 
\end{tabular}
\end{table}

\section{Qualitative Results}

\subsection{The HOZ graph visualization}

Figure \ref{fig:Scene-Zone-Graph} illustrates the visualization of
our HOZ graph. We visualize the zones nodes in a scene-wise HOZ graph
(e.g., living room), which is the fusion of 20 room-wise HOZ
graphs. There are 8 zones marked with different colors and each zone
consists of similar objects distribution. Even though there are overlapped
objects among zones, each zone has semantically representative objects.  For instance, in Figure \ref{fig:Scene-Zone-Graph},
$zone_{2}$, $zone_{3}$, $zone_{6}$ focus on laptop, garbage can
and television, respectively.

\subsection{Navigation trajectory}

Figure \ref{fig:Visualization} qualitatively compares our method
with the baseline in RoboTHOR. Benefiting from the sub-goals guidance
and online-updating of proposed HOZ graph, agent can still adopt reasonable
actions even in the long trajectory unseen navigation task, while
the baseline model often falls into confusion and struggles with spinning
around.

\end{document}